\documentclass[letterpaper,10pt,conference]{ieeeconf}
\IEEEoverridecommandlockouts       
\usepackage{amssymb}
\usepackage{amsmath}
\usepackage[english]{babel}
\usepackage{float}
\usepackage{graphicx}
\usepackage{hyperref}
\usepackage{mathtools}

\usepackage{filecontents}
\usepackage[noadjust]{cite}
\usepackage[font=footnotesize,labelfont=footnotesize]{subcaption}
\usepackage[font=footnotesize,labelfont=footnotesize]{caption}
\usepackage{comment}
\usepackage{authblk}
\usepackage{multirow}
\usepackage{xcolor}
\usepackage{algorithm}
\usepackage{algorithmic}
\usepackage{xspace}
\usepackage{graphicx}

\definecolor{darkblue}{rgb}{0.15,0.15,0.55}
\definecolor{lightgrey}{rgb}{0.75,0.75,0.75}

\newcommand{\nphard}{$\mathcal{NP}$-hard\xspace}

\newcommand{\thm}{\noindent \textbf{Theorem}\xspace}

\newcommand{\pf}{\noindent \textbf{Proof}\xspace}

\newcommand{\qed}{\hfill $\square$}

\begin{document}
\title{\LARGE \bf Coordination of two robotic manipulators for object retrieval in clutter}
\author{Jeeho Ahn$^1$, ChangHwan Kim$^1$, and Changjoo Nam$^{2,*}$
\thanks{This work was supported in part by the Technology Innovation Program and Industrial Strategic Technology Development Program (10077538, Development of manipulation technologies in social contexts for human-care service robots) and the Institute of Information \& communications Technology Planning \& Evaluation (IITP) grant funded by the Korea government (MSIT) (2020-0-01389, Artificial Intelligence Convergence Research Center at Inha University).
$^1$Korea Institute of Science and Technology, $^2$Inha University.
$^*$Corresponding author: {\tt\small cjnam@inha.ac.kr}. }\\
}

\maketitle

\begin{abstract}
We consider the problem of retrieving a target object from a confined space by two robotic manipulators where overhand grasps are not allowed. If other movable obstacles occlude the target, more than one object should be relocated to clear the path to reach the target object. With two robots, the relocation could be done efficiently by simultaneously performing relocation tasks. However, the precedence constraint between the tasks (e.g, some objects at the front should be removed to manipulate the objects in the back) makes the simultaneous task execution difficult.

We propose a coordination method that determines \textit{which robot relocates which object}  so as to perform tasks simultaneously. Given a set of objects to be relocated, the objective is to maximize the number of turn-takings of the robots in performing relocation tasks. Thus, one robot can pick an object in the clutter while the other robot places an object in hand to the outside of the clutter. However, the object to be relocated may not be accessible to all robots, so taking turns could not always be achieved. Our method is based on the optimal uniform-cost search so the number of turn-takings is proven to be maximized. We also propose a greedy variant whose computation time is shorter. From experiments, we show that our method reduces the completion time of the mission by at least 22.9\% (at most 27.3\%) compared to the methods with no consideration of turn-taking.

\end{abstract}

\section{Introduction}
\vspace{-3pt}

% Manipulation planning in clutter, hardness
Manipulation planning to rearrange objects in clutter has recently received significant attention~\cite{qureshi2021nerp, saleem2020planning,han2018complexity,bejjani2021learning} as it has a wide variety of applications from domestic services to manufacturing. Among them, there has been a line of research~\cite{wang2021uniform,nam2020fast,cheong2020where,ahn2021integrated,han2018efficient,lee2019efficient} that aims to retrieve a target object from cluttered and confined spaces where overhand grasps are not allowed, like shelves and fridges. These works focus on generating plans for what to relocate in what order. Since multiple objects are manipulated, distributing the manipulation tasks to multiple robots may enable faster completion of the mission.

\begin{figure}[t]
%\vspace{-5pt}
\captionsetup{skip=0pt}
    \captionsetup{skip=0pt}
    \centering
   \begin{subfigure}{0.24\textwidth}
   \captionsetup{skip=0pt}
	\includegraphics[width=\textwidth]{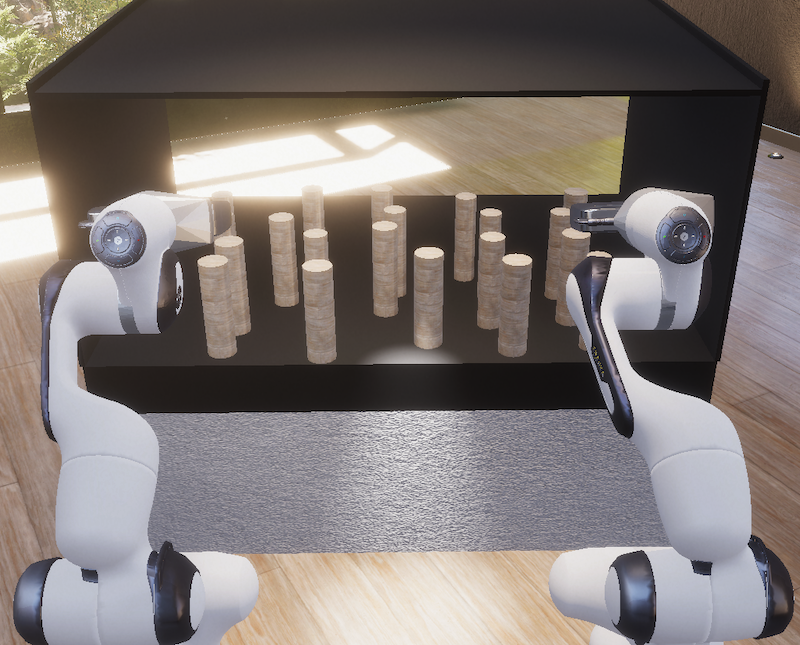}
	\caption{A confined and cluttered space}
    \label{fig:ex_a}
  \end{subfigure}
  \begin{subfigure}{0.225\textwidth}
  \captionsetup{skip=0pt}
	\includegraphics[width=\textwidth]{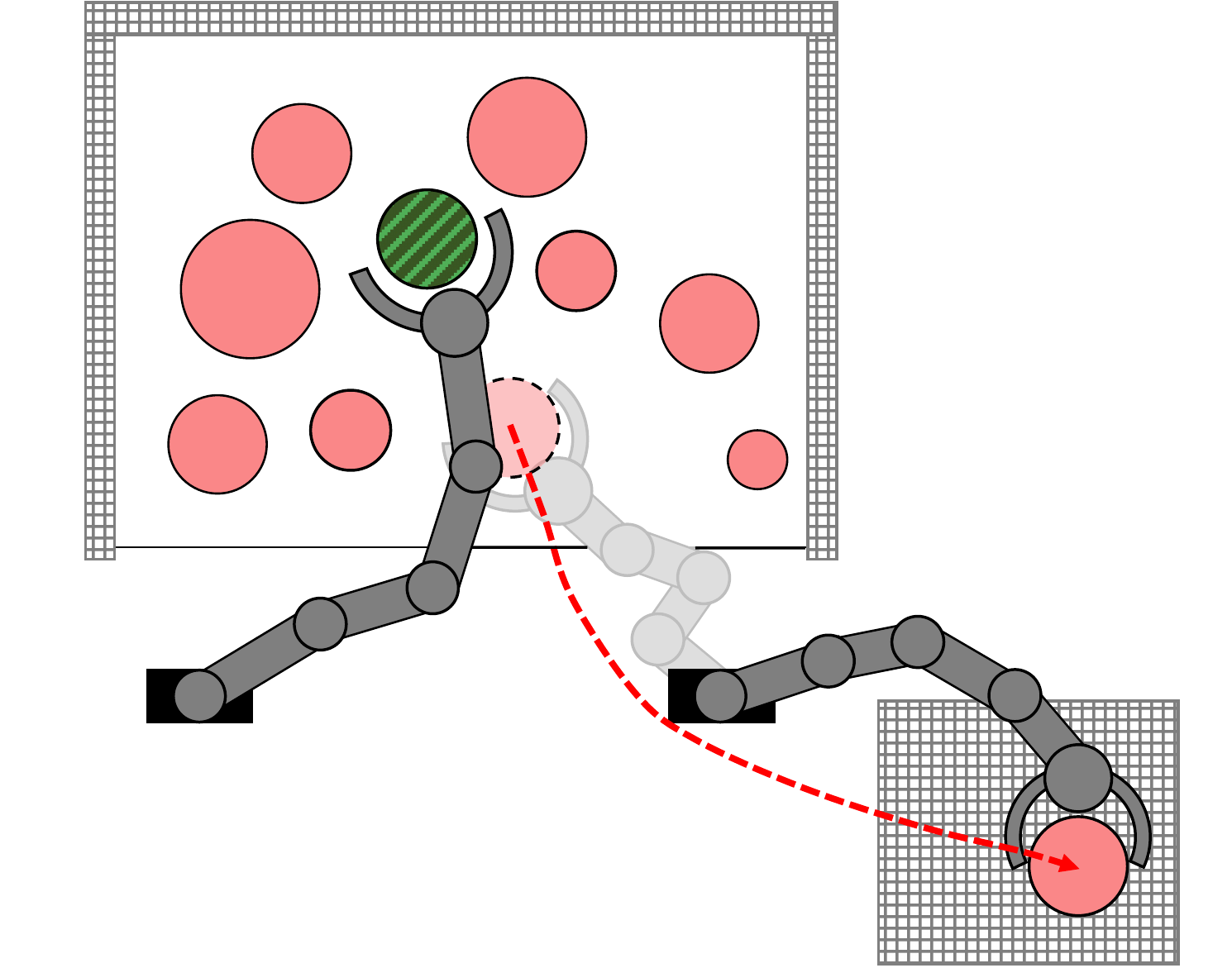}
	\caption{A simultaneous execution of tasks}
    \label{fig:ex_b}
  \end{subfigure}
  \caption{Coordination of two robots for object retrieval. (a) An example configuration of the robots and objects. (b) While the right robot is placing an object, the left robot can pick an object that becomes accessible after the right robot removes the object being placed.}
  \label{fig:ex}\vspace{-10pt}
\end{figure}

% Benefit of using multiple arms for the manipulation in clutter (multiple objects to be manipulated), alternation
However, coordinating multiple manipulators in a confined space (Fig.~\ref{fig:ex_a}) is challenging as the robots have limited (and varying) configuration spaces if they move simultaneously. The manipulation tasks have precedence constraints since some objects only can be manipulated after some front objects are relocated. This constraint and the limited configuration space may force only one robot to execute a task at a time. In order for the robots to work simultaneously, the actions of the robots need to be well coordinated. For example, one robot picks an object while the other places an object outside the clutter (Fig.~\ref{fig:ex_b}). Then the robots can comply with the precedence constraints and use a larger configuration space. However, such coordination is not possible if only one of the robots can access the objects (e.g., the path from the other robot to the objects is blocked).

%\begin{figure}
%\centering     %%% not \center
%\subfigure[A confined %space]{\label{fig:a}\includegraphics[width=42mm]{figures/first_view_unity_cropped.png}}
%\subfigure[A simultaneous execution of tasks]{\label{fig:b}\includegraphics[width=42mm]{figures/ex2.pdf}}
%\caption{Coordination of two robots for object retrieval. (a) An example configuration of the robots and objects. (b) While the right robot is placing an object, the left robot can pick an object that becomes accessible after the right robot removes the object being placed.}
%\end{figure}

%Note that we limit the scope of this work to two manipulators and pick-and-place actions and show the feasibility of extensions to more than two manipulators and nonprehensile actions.

% Objective: minimize? reduce the total  execution time
% Approach: achieve the objective by finding a sequence of object-arm allocation until grasping the target after relocating all obstacles
% Detail: (specifically) t-graph, two arms' relocation plan, A* search, cost (admissible heuristic), replanning, extension to multi-arm, proofs for completeness, optimality, first work

Therefore, we develop a method that coordinates robot
actions while considering the accessibility of the objects.
We aim to maximize the number of turn-takings of the
robots in performing relocation tasks. A relocation task
is defined as picking an object and placing it in another
location. If two robots take turns, they are likely to perform
tasks simultaneously so the retrieval mission can be done
efficiently. Specifically, the method finds relocation tasks
and allocates them to the robots using the uniform-cost search.
The search finds an optimal allocation describing \textit{what to relocate in what order using which robot}
such that the number of turn-takings is maximized. Given the allocation,
the method sequences robot actions (e.g., pick, place) for
simultaneous execution. Once the method finds collision-free
motions for all actions, the robots start relocation.

%During execution, either of the robots may fail to perform a task owing to dynamic situations (e.g., object positions change from unexpected collisions). Then the method modifies the relocation plan by updating the states of the objects and run the search again.

However, the retrieval problem is computationally intractable even with only one robot~\cite{nam2021fast}. Although the method can replan if some of the motions turn out to be inexecutable, repeated replanning could delay the completion of the mission. To prevent such delays, we develop a greedy variant that determines the allocation in an online manner. 

The following are contribution of this work:
\begin{itemize}
    \item We propose a method that finds an allocation of relocation tasks to robots with the maximum number of turn-takings. Through sequencing actions to perform tasks simultaneously, the mission can be completed quickly.
    \item We also propose a greedy variant for situations where replanning occurs repeatedly. 
    \item We provide analyses of the methods regarding optimality, time complexity, and completeness.
    \item We show the results from extensive experiments and comparisons with other methods that do not set out to optimize the number of turn-takings.
\end{itemize}
%%%%%% revised up to here 2021/09/13 2:13AM

\section{Related Work}
\vspace{-3pt}

% focusing on hand off 
Planning for multiple robotic arms has been studied in many contexts. One of the topics that received attention is hand-off between robots to transfer objects. \cite{umay2019integrated,cohen2015planning,shome2019anytime,shome2020synchronized} consider task and/or motion planning for multiple manipulators to transfer a single object between two locations via hand-offs and find collision-free motions for the transfer. While these approaches can be helpful for our object retrieval to transfer picked objects to the place outside the clutter, they do not directly solve the retrieval problem. 

%Also, they consider transferring a single object only so cannot provide solutions to the rearrangement problem of multiple objects.

% no precedence constraint
%Some other work consider manipulation of multiple objects. In~\cite{shome2020synchronized}, the multiple manipulators transfer multiple objects via hand-offs. 
In~\cite{sepulveda2020robotic}, a planning and control method for dual-arm fruit harvesting is presented where the robot is in front of multiple fruits to be harvested. The arms of the robot can harvest fruits (aubergine) simultaneously, in turn, or in collaboration with each other. \cite{shome2021fast} use two arms to rearrange blocks on a table to a predefined configuration. Since these works consider the arrangement of objects where any object can be picked without relocating some others, the manipulation tasks do not have a precedence constraint.  
\cite{kimmel2016scheduling} propose methods based on precomputed information about collisions of two arms, called the coordination matrix encoding all trajectories of the arms to perform all manipulation tasks. They determine the order of the objects to be manipulated and find the motions to pick and place the objects using the matrix. The tasks in this work do not have precedence constraints, so the two arms can manipulate objects simultaneously. Also, the methods take a long time to find a solution with clutter environments, which ranges from several minutes to several tens of minutes (the time varies whether the matrix is precomputed or not).

%\cite{suarez2018interleaving} consider the collaboration of two robot arms to solve a puzzle. One robot holds a toy sphere with cavities of different shapes. Among several puzzle pieces distributed on a table, the other robot picks the pieces (one at a time) to insert them into the sphere. 

% precedence but not clutter
The methods proposed in~\cite{behrens2020simultaneous,behrens2019constraint}  integrate task allocation and motion scheduling for task optimization for dual-arm coordination. They are with a solver that finds solutions for constraint satisfaction problems so they can handle precedence constraints. Despite being generic and computationally efficient, they consider environments where robots do not have cramped configuration spaces (like an open space but not a clutter).
%Also, it uses a precomputed roadmap for the efficient online computation of collision-free motions. In our problem, precomputing such a roadmap is not appropriate as the configuration space in our environment changes after every relocation action.
In~\cite{kabir2020incorporating}, the proposed method uses a surrogate checker to evaluate the feasibility of task assignments (i.e., if collision-free motions exist) to work in an environment that is more constrained. Also, the method can handle tasks with precedence constraints. However, it requires a predefined task network specifying tasks with pre-conditions and post-conditions. We do not want to hand-code such conditions (i.e., what to relocate in what order in our problem) but find using a geometric reasoner.

% First work that considers all of the following
More importantly, these works do not directly solve the object retrieval problem. To the best of our knowledge, our work is the first to tackle the coordination of multiple manipulators for object retrieval in clutter (so motion planning is nontrivial) where manipulation tasks have precedence constraints. This work aims to fill the gap by developing fast methods to determine relocation tasks with precedence constraints and allocate them to two robots.

\section{Problem Description}
\label{sec:prob}
\vspace{-3pt}

In order to retrieve a target object from clutter using two manipulators, there must be several processes to be completed, such as sensing, perception, task planning and allocation, motion planning for the allocated tasks, grasp planning, and control. Among these, we focus on task planning and allocation and motion planning to distribute relocation tasks. The target retrieval problem among multiple movable objects has shown to be computationally intractable with one manipulator~\cite{nam2021fast}. Since adding more robots increases the complexity of the problem structure, the target retrieval with more than one robot must be a more complicated problem.

\subsection{Assumptions}
\vspace{-3pt}

We have several assumptions to focus on the problems that have not been tackled in the literature: (i) Each robot can manipulate one object at a time, and each object requires exactly one robot to manipulate it.\footnote{According to the taxonomy for multi-robot task allocation~\cite{gerkey2004formal}, our problem falls into the ST-SR-TA category (\textbf{S}ingle-\textbf{T}ask robots, \textbf{S}ingle-\textbf{R}obot tasks, \textbf{T}ime-extended \textbf{A}ssignment), which is strongly \nphard~\cite{bruno1974scheduling}.} (ii) The computations for task planning, task allocation, and motion planning are centralized. (iii) The trivial problem instances where the robot can grasp the target without relocating any object are not considered. The infeasible problem instances where none of the objects can be grasped are not considered. (iv) The robots use prehensile actions only (i.e., pick, place).   (v) Overhand grasps are not allowed (e.g., the top is blocked by shelves). (vi) Objects are modeled by 3D cylinders (which could have different radii) so the objects can be grasped from any direction. We will relax some of these assumptions in our future work once we solve the coordination problem.\footnote{Our previous work can be employed for nonprehensile actions~\cite{lee2021tree} and varying reachable directions~\cite{nam2020fast}.}

\subsection{Problem definition}
\vspace{-3pt}

Our goal is to coordinate two manipulators efficiently to retrieve the target object from the confined space cluttered with other movable obstacles. To achieve this goal, we need to solve three subproblems: (A) determine the objects to relocate to clear the path to the target and the order of their relocation (i.e., finding the relocation tasks and their precedence constraints); (B) allocate the relocation tasks to the robots; and (C) find collision-free motions of the robots for the allocated tasks. Since the entire problem is computationally intractable, we solve them sequentially for faster computation. Thus, each subproblem is conditioned on the result of the previous subproblem. 

We set objective values for the subproblems. For (A), we aim to minimize the number of relocated objects, which has shown to be effective in reducing the mission completion time~\cite{yu2021rearrangement,nam2021fast}. For (B), the objective is to maximize the number of turn-takings of the robots in performing relocation tasks. Lastly, we aim to minimize the execution time of the relocation tasks, known as makespan. 

Suppose that an environment is with $N \in \mathbb{Z}^{\ge 2}$ objects and $M  \in \mathbb{Z}^{+}$ robotic manipulators where we only consider $M=2$ in this work. The set  $\mathcal{O} = \{o_1, \cdots, o_{N-1}, o_t\}$ includes all objects with $N-1$ movable obstacles and one target object $o_t$. Likewise, $\mathcal{R} = \{r_1, \cdots, r_M\}$ includes all robots. The centroid and radius of an object $i$ is described by $(x_i, y_i, z_i)$ and $r_i$, respectively. The target is within the intersection of the workspaces of the robots so $(x_t, y_t, z_t) \in \mathcal{W} = \mathcal{W}_1 \cap \cdots \cap \mathcal{W}_M \neq \emptyset$ for all $i = 1, \cdots, N$. 

We define three action primitives of the robots to perform relocation tasks: \texttt{pick}, \texttt{place}, \texttt{standby}. The \texttt{pick} and \texttt{place} actions are specified by a robot ID and a Cartesian coordinate $p \in \mathbb{R}^3$ to indicate the goal location  to perform the actions. $\texttt{standby}(r_i)$ moves $r_i$ to a predefined standby pose. For example, $\texttt{pick}(r_1, p_1)$ let $r_1$ move to $p_1$, and grasp the object at $p_1$. The action $\texttt{place}(r_1, p_2)$ indicates that $r_1$ moves to $p_2$, and releases the object in hand. 

%%%%%% revised up to here 2021/09/13 11:02AM

\begin{figure}
\vspace{-5pt}
\captionsetup{skip=0pt}
    \captionsetup{skip=0pt}
    \centering
   \begin{subfigure}{0.2\textwidth}
   \captionsetup{skip=0pt}
	\includegraphics[width=\textwidth]{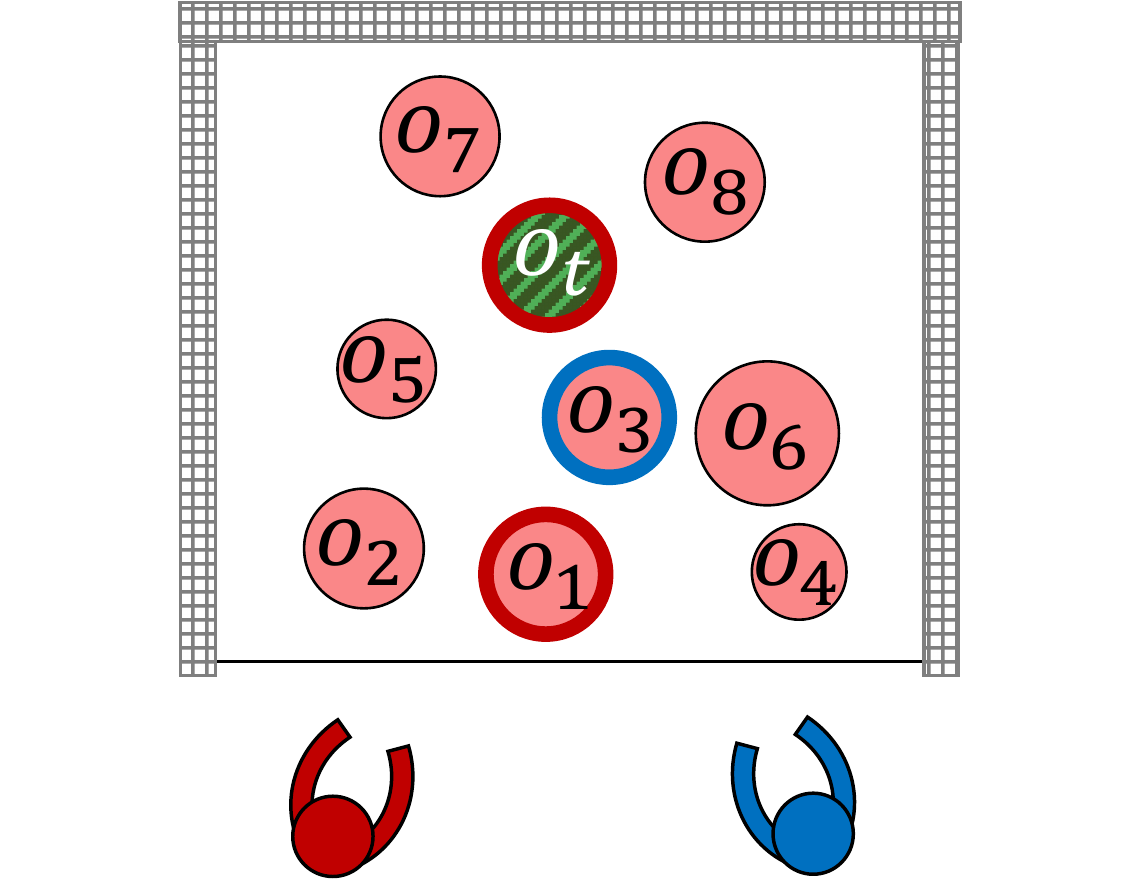}
	\caption{$\mathcal{X}_R = (r_1, r_2, r_1)$}
    \label{fig:alloc_a}
  \end{subfigure}\quad
  \begin{subfigure}{0.195\textwidth}
  \captionsetup{skip=0pt}
	\includegraphics[width=\textwidth]{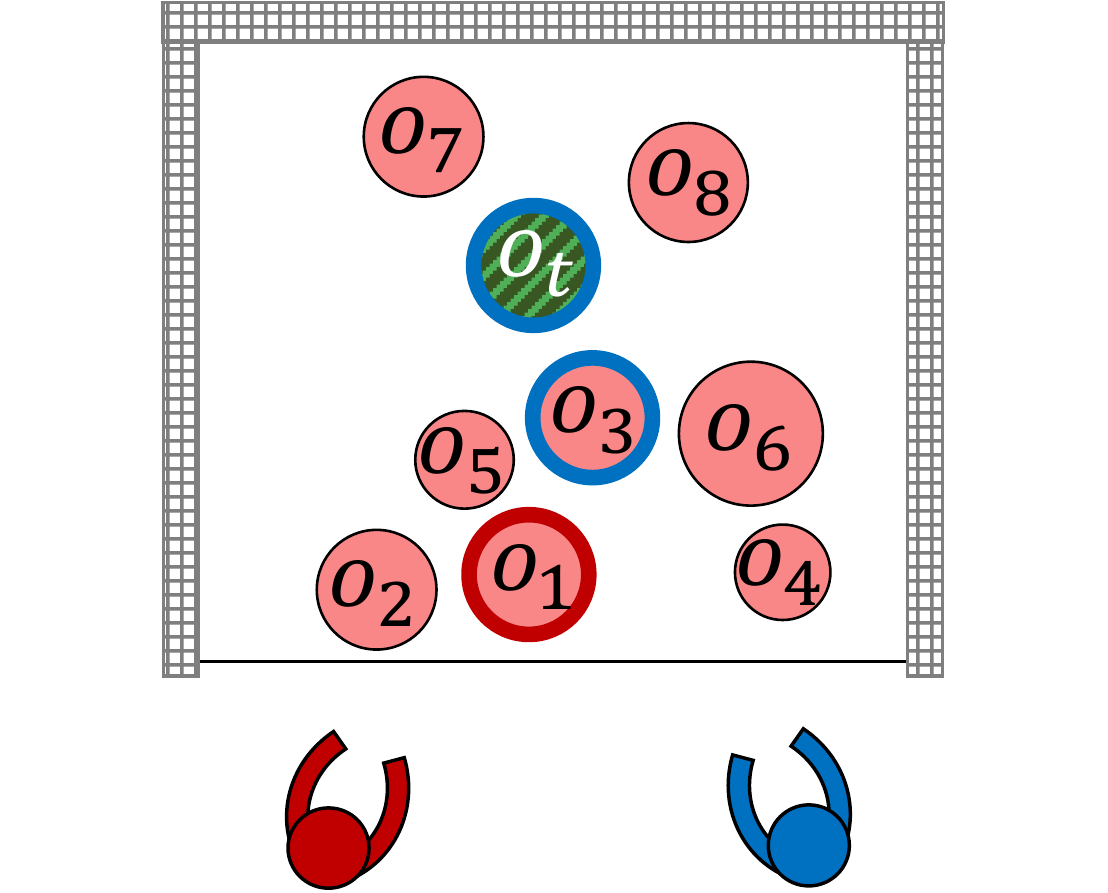}
	\caption{$\mathcal{X}_R = (r_1, r_2, r_2)$}
    \label{fig:alloc_b}
  \end{subfigure}
  \caption{Example alloctions for $\mathcal{O}_R = (o_1, o_3, o_t)$. In (b), $r_1$ cannot access $o_t$ even after $o_1$ and $o_3$ are removed because $o_2$ and $o_5$ occlude $o_t$.}
  \label{fig:alloc}\vspace{-15pt}
\end{figure}

\noindent \textbf{(A) Object rearrangement planning (ORP)}: Let $\mathcal{O}_R$ be a tuple representing the sequence of objects to be relocated (including the target) where $|\mathcal{O}_R| = k \le N$. A mathematical definition of the ORP is to find $\mathcal{O}_R$ that minimizes $k$. An example sequence $\mathcal{O}_R = (o_1, o_3, o_t)$ means that $o_1$ and $o_3$ should be removed sequentially in order to retrieve the target, as shown in the both instances in Fig.~\ref{fig:alloc}.

\noindent \textbf{(B) Multi-manipulator task allocation (MMTA):} Let $\mathcal{X}_R$ be a tuple representing the sequence of robots to relocate objects in $\mathcal{O}_R$. In other words, $\mathcal{X}_R$ is an allocation of relocation tasks to robots. Suppose that we have $\mathcal{O}_R = (o_1, o_3, o_t)$ and $\mathcal{X}_R = (r_1, r_2, r_1)$ (see the color coding in Fig.~\ref{fig:alloc_a}). The red $r_1$ and blue $r_2$ relocate $o_1$ and $o_3$, respectively. Then $r_1$ finally retrieves $o_t$. Let $t$ be the number of turn-takings of the robots in $\mathcal{R}$ to relocate $\mathcal{O}_R$ where $0 \le t \le k-1$. The MMTA is to find $\mathcal{X}_R$ such that $t$ is maximized. An object $o_i$ may not be relocated by $r_i$ if $r_i$ cannot access $o_i$ owing to occlusions. Thus, the robots may not always be able to take turns like $r_1$ (red) in Fig.~\ref{fig:alloc_b} that cannot access $o_t$ even after $o_1$ and $o_3$ are removed owing to $o_2$ and $o_5$. Thus, $\mathcal{X}_R = (r_1, r_2, r_2)$.

\begin{figure}[t]
\vspace{-3pt}
\captionsetup{skip=0pt}
    \captionsetup{skip=0pt}
    \centering
   \begin{subfigure}{0.46\textwidth}
   \captionsetup{skip=0pt}
	\includegraphics[width=\textwidth]{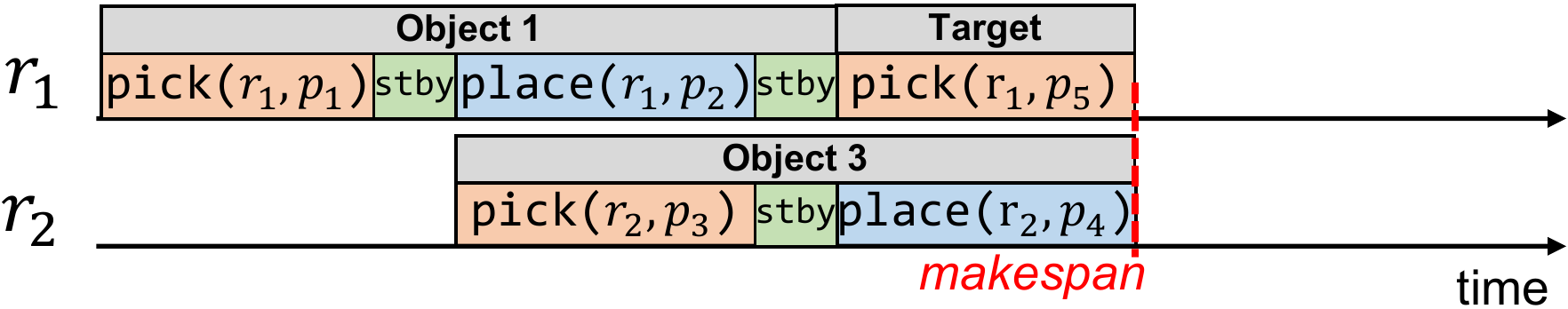}\vspace{-2pt}
	\caption{$\mathcal{O}_R = (o_1, o_3, o_t)$ and $\mathcal{X}_R = (r_1, r_2, r_1)$}
    \label{fig:makespan_b}
  \end{subfigure}
  \begin{subfigure}{0.46\textwidth}
  \captionsetup{skip=0pt}
	\includegraphics[width=\textwidth]{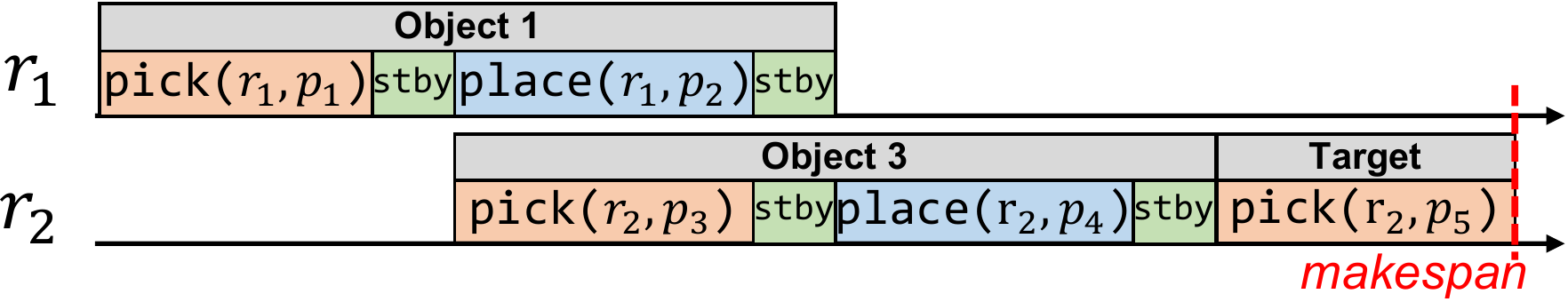}\vspace{-2pt}
	\caption{$\mathcal{O}_R = (o_1, o_3, o_t)$ and $\mathcal{X}_R = (r_1, r_2, r_2)$}
    \label{fig:makespan_c}
  \end{subfigure}
  \caption{Minimum makespan action sequencing for the examples in Fig.~\ref{fig:alloc}}
  \label{fig:makespan}\vspace{-3pt}
\end{figure}

\noindent \textbf{(C) Minimum makespan action sequencing (MMAS):} Minimum makespan motion planning (MMMP) is to minimize the total completion time (i.e., makespan) to relocate $\mathcal{O}_R$ by $\mathcal{X}_R$. Since the MMMP on a grid (i.e., in the discrete space) is strongly \nphard~\cite{demaine2019coordinated}, MMMP in the continuous space of robot joints is also \nphard. To develop a practical method that works fast, we simplify the difficult problem to an action sequencing problem (MMAS). We let the robots perform $\texttt{pick}(r_i, o_a)$ and $\texttt{place}(r_j, o_b)$ simultaneously if $i \neq j$ and $a \neq b$. If the robots execute the actions simultaneously, we synchronize the start time of the actions (their end time does not necessarily synchronize). \texttt{standby} actions are inserted between every \texttt{pick} and \texttt{place} to start the following action from the same pose (necessary for upfront offline motion planning). This synchronization makes the problem easy since we do not have to be concerned about allocating the actions to specific time frames. The action sequences of the examples from Fig.~\ref{fig:alloc} are shown in Fig.~\ref{fig:makespan}.

%Goal: alternating the arms as many as possible
%1. determining what to relocate in what order ()
%2. allocating the tasks to the arms considering the feasibility of having collision-free motions (using a surrogate checker, VFH+)
%3. sequencing motions based on the task allocation

\section{The Methods for Coordinating Robots}
\vspace{-3pt}

In this section, we describe the methods for the three subproblems (A--C). For (A), we introduce an existing method that finds a relocation plan $\mathcal{O}_R$. Then we propose our methods for (B) and (C). One method for (B) uses the optimal uniform-cost search with exponential time complexity in the number of objects. Thus, we also propose a greedy variant that finds the object to relocate and the robot to perform relocation in an online fashion.

 %to allocate relocation tasks to the robots and find action sequences of the robots to generate coordinated motion plans

\subsection{Preliminary: Object rearrangement planning}
\vspace{-3pt}

\begin{figure}
\captionsetup{skip=0pt}
    \captionsetup{skip=0pt}
    \centering
   \begin{subfigure}{0.2\textwidth}
   \captionsetup{skip=0pt}
	\includegraphics[width=\textwidth]{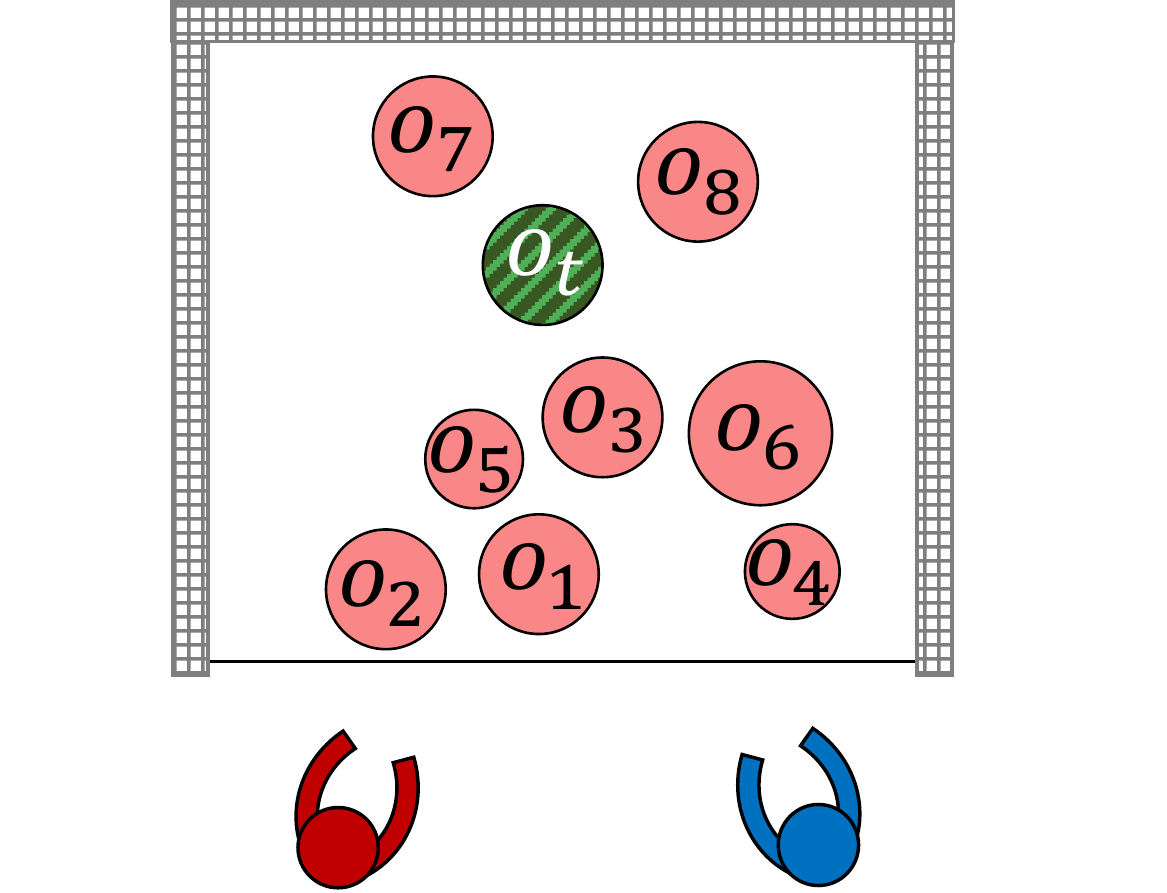}
	\caption{An instance with two robots}
    \label{fig:scene_for_graph}
  \end{subfigure}
  \begin{subfigure}{0.26\textwidth}
  \captionsetup{skip=0pt}
	\includegraphics[width=\textwidth]{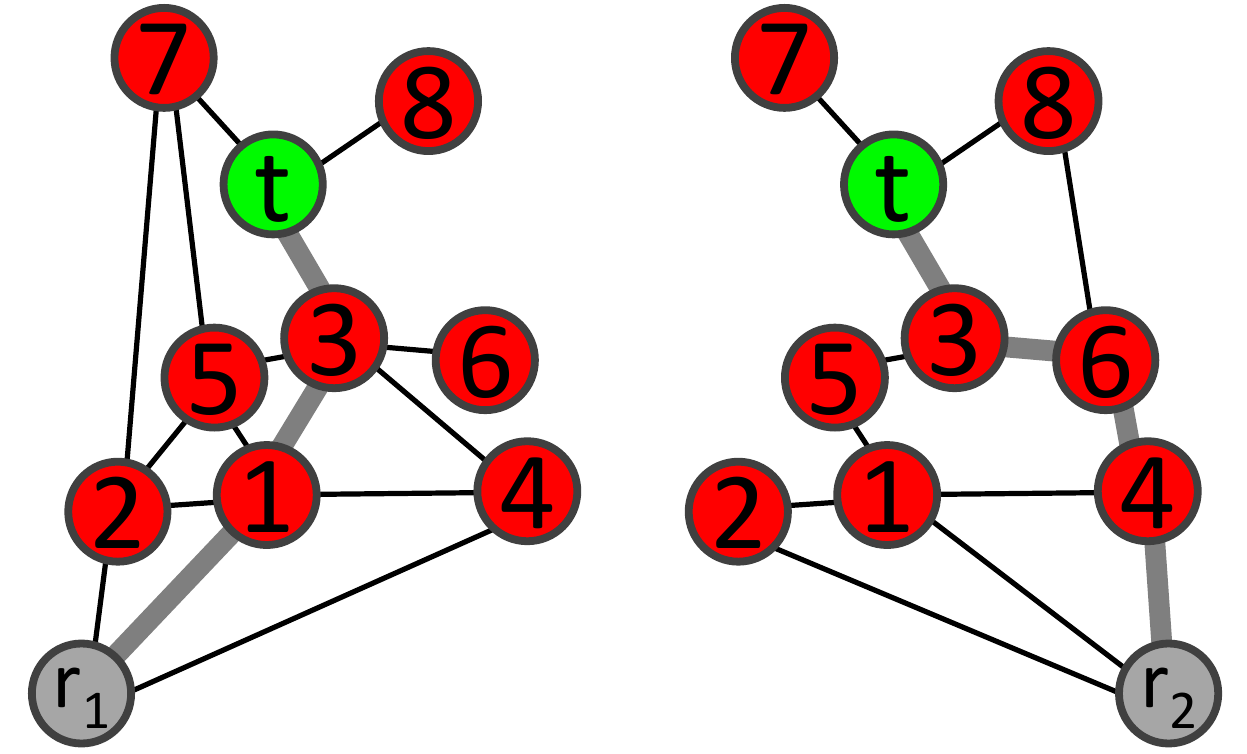}
	\caption{T-graphs and the paths in the graphs}
    \label{fig:graph}
  \end{subfigure}
  \caption{Example T-graphs. The relocation plans for $r_1$ and $r_2$ are $\mathcal{O}^1_R = (o_1, o_3, o_t)$ and $\mathcal{O}^2_R = (o_4, o_6, o_3, o_t)$, respectively.}
  \label{fig:graphs}\vspace{-10pt}
\end{figure}

We employ a method proposed in~\cite{nam2020fast,nam2021fast} to find $\mathcal{O}_R$. This method generates a Traversability graph (T-graph). The graph represents movable paths of objects in clutter. A node represents the location of an object or the end-effector. An edge is connected between a pair of nodes if a collision-free path of the end-effector exists between the locations corresponding to the nodes. Collisions are examined by a fast surrogate checker~\cite{lee2019efficient} without finding motions of the whole arm. Motion planning is not necessary with this checker, so the graph construction runs in polynomial time.
Then a shortest path in the graph from the robot node to the target node is a sequence of nodes representing the smallest number of objects to relocate. The two robots could have different T-graphs and shortest paths to the target. Example graphs and paths (the nodes on the bold gray edges) for $r_1$ and $r_2$ are shown in Fig.~\ref{fig:graph}. The relocation plans for $r_1$ and $r_2$ are $\mathcal{O}^1_R = (o_1, o_3, o_t)$ and $\mathcal{O}^2_R = (o_4, o_6, o_3, o_t)$, respectively.

% since the accessible objects by the robots vary depending on the location of the robots and the directions that the robots approach the objects

\subsection{Multi-manipulator task allocation}
\vspace{-3pt}

Based on $\mathcal{O}^i_R$ for $i = 1, 2$ from (A), we compute the allocation of the relocation tasks to $r_i$. We propose two methods where \textsc{Search} finds the allocation for all tasks while \textsc{Greedy} instantaneously determines the robot to relocate. 

\begin{figure*}
\captionsetup{skip=0pt}
    \centering
    \includegraphics[width=0.95\textwidth]{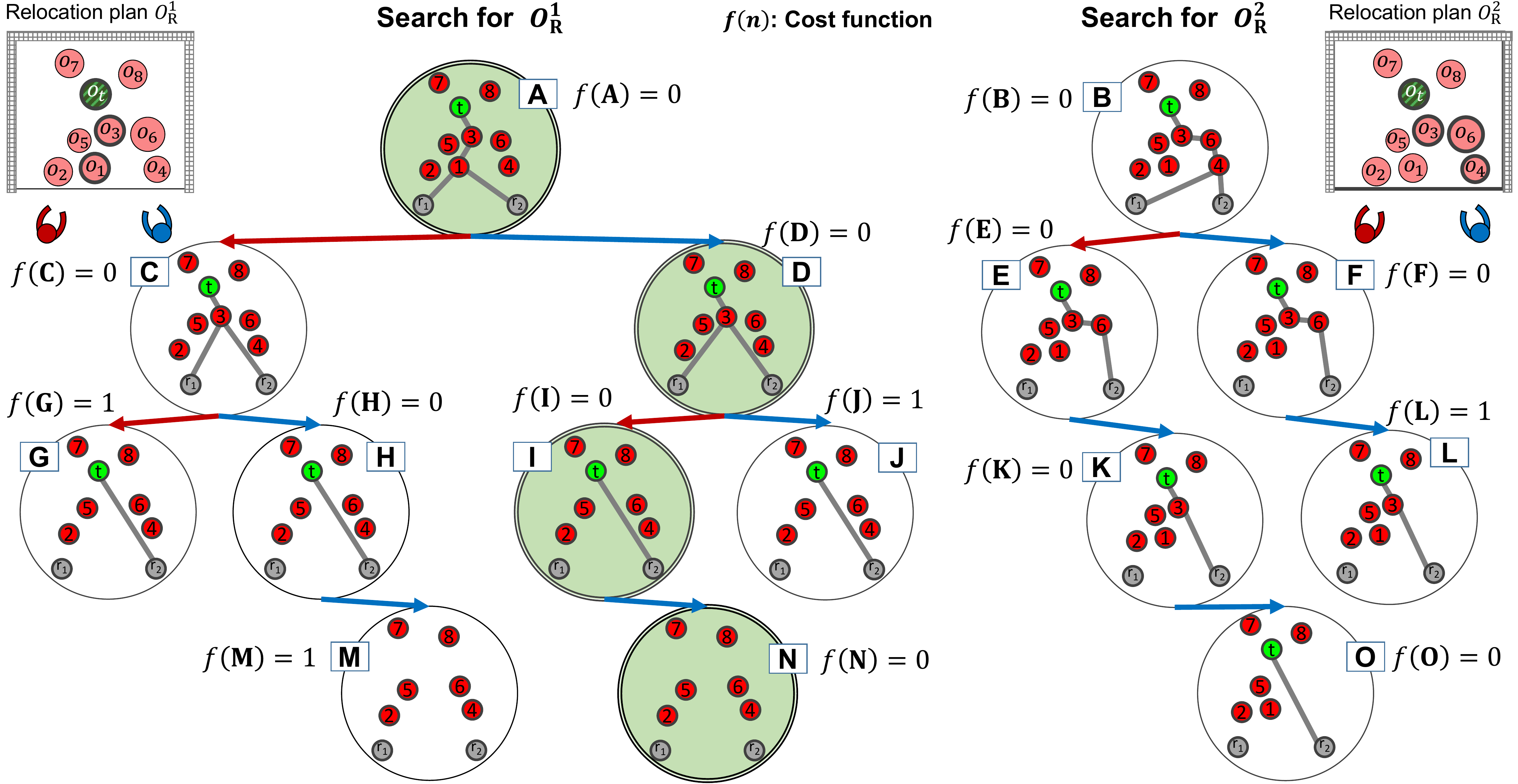}
    \caption{An example result of \textsc{Search}. A simplified T-graph is inscribed in each node. From the root node (not shown here due to the space limit), two children \textsf{A} and \textsf{B} are generated for $\mathcal{O}^1_R$ and $\mathcal{O}^2_R$. \textsf{A} is expanded as it is generated earlier. Nodes \textsf{C} and \textsf{D} are generated. Then \textsf{B}, \textsf{C}, and \textsf{D} with the same cost are on the frontier. The oldest \textsf{B} is chosen for expansion and \textsf{E} and \textsf{F} are generated. Then \textsf{C} is expanded so \textsf{D}, \textsf{E}, \textsf{F}, \textsf{G}, and \textsf{H} are in the frontier list. The oldest \textsf{D} among those with the minimum cost is chosen then now \textsf{I} and \textsf{J} are added to the frontier list. Among the nodes with zero cost, \textsf{E} and \textsf{F} are expanded in turn so \textsf{K} and \textsf{L} are generated. In \textsf{E} and \textsf{F}, $r_1$ cannot access the next object (no edge in the T-graph between the robot node and the object node) so only the right subtrees are generated. Next, \textsf{M} is generated after \textsf{H} expands. The target is removed in \textsf{M}, but the goal check is done after the node is chosen for expansion. Next, \textsf{I} is expanded and \textsf{N} is generated. At this point, the frontier nodes with zero cost are \textsf{K} and \textsf{N}. After \textsf{O} is generated, \textsf{N} is chosen for expansion. The goal check finds the goal state \textsf{N} so the search terminates.}
    \label{fig:search}\vspace{-10pt}
\end{figure*}

\subsubsection{Optimal search}
\label{sec:search}
%Each node corresponds to a state of objects before or after performing a relocation task.

Given $\mathcal{O}^i_R$ and $r_i$ for $i = 1, 2$, \textsc{Search} performs the uniform-cost search. Two initial nodes are generated from the root for $\mathcal{O}^1_R$ and $\mathcal{O}^2_R$, respectively (the root node is omitted in Fig. 5 to save the space). As a visual guide, we draw the T-graphs inside the nodes only with the path to the target node for simplicity. Since the two initial nodes, a child node represents the state of objects after one object is relocated. The child in the left (via the red arrow) represents the state where the object is relocated by $r_1$ (the red robot). For example, node \textsf{C} in Fig.~\ref{fig:search} represents the state where $o_1$ is relocated by $r_1$ from the state in \textsf{A}. Similarly, the blue arrow means that the relocation is done by $r_2$ (the blue robot). The search terminates if the goal state, where the target is finally relocated, is found. The nodes highlighted by green in Fig.~\ref{fig:search} represent the solution, that is $\mathcal{X}_R = (r_2, r_1, r_2)$ for $\mathcal{O}_R = (o_1, o_3, o_t)$.

The plans $\mathcal{O}^1_R$ and $\mathcal{O}^2_R$ have the smallest numbers of objects to be relocated by $r_1$ and $r_2$, respectively. Although only one search can run for either of the plans with fewer tasks, like $\mathcal{O}^1_R$, \textsc{Search} runs the search for both plans to ensure that the solution relocates the smallest number of objects even if motion planning fails. Suppose that we expand only the left subtree of the root. If no collision-free motion is found for relocating any of the objects in $\mathcal{O}^1_R$, a new path to the target is computed after updating the T-graph. This new path could result in relocating more objects in total than executing the plan $\mathcal{O}^2_R$. If so, $\mathcal{O}^2_R$ should be chosen in order to maintain the least number of objects to be relocated.

%Note that replanning never produces a plan with a smaller number of tasks than the initial plan since the updated T-graph has less edges than the graph used to compute the initial plan (but a tie can be produced).

The uniform-cost search is guided by the evaluation function $f(n) = g(n)$ for node $n$ where $g(n)$ represents the sum of penalties imposed if no turn-taking occurs. Nodes have zero cost up to depth 2 of the tree (from the root to its grandchildren) since performing no or a single task cannot incur any turn-taking. From depth 3 (from \textsf{G} in Fig.~\ref{fig:search}), a unit penalty is given if the robots do not take turns. For example, \textsf{G} in Fig.~\ref{fig:search} has $g(\textsf{G}) = 1$ since no turn-taking occurs between \textsf{C} and \textsf{G}. The node for expansion is chosen such that the chosen node has the minimum $f(n)$ among the nodes in the frontier list. If some nodes on the frontier have the same cost, the node generated earlier is chosen. In Fig.~\ref{fig:search}, the nodes are named alphabetically according to the order that they are generated. Since the behavior of the search is the same as the traditional uniform-cost search~\cite{russell2002artificial} and we have the space limit, we do not provide the pseudocode but a detailed illustration in Fig.~\ref{fig:search} instead.

Once the search is done, the tasks and their allocation $\mathcal{O}_R$ and $\mathcal{X}_R$ can be obtained from the nodes chosen for expansion. This allocation is then forwarded to the next stage (MMAS) for action sequencing, motion planning, and execution. However, motion planning could fail to pick or place some of the objects in the plan. Then, the process returns to the ORP and MMTA to replan since the task planning and allocation turn out to be invalid at runtime. Replanning goes through the same procedure described above but uses an updated T-graph where the edge between the robots and the failed object is invalidated.

\subsubsection{Greedy version}
\label{sec:greedy}

\textsc{Search} performs an upfront computation to find the allocation for all tasks. If all tasks have feasible motions so that no replanning occurs, few tens of seconds of computation time can be acceptable. If motion planning fails for the allocated tasks, another few tens of seconds for replanning is necessary. Thus, we develop an online greedy version for cases where fast planning is paramount.

Given $\mathcal{O}_R$, \textsc{Greedy} chooses the robot performing a task instantaneously in each turn. The choice is made based on the penalty used in \textsc{Search}. Unlike \textsc{Search} summing up all penalties up to the current node, \textsc{Greedy} looks if the turn-taking occurs in this turn. There is no notion of turn-taking when performing the first task, so the robot closer to the object is chosen. Afterward, the robot that did not perform the relocation task in the previous turn relocates the next object. If only one robot can access the object, it should be chosen. If no robot can access the object, the distances between all pairs of the robot and accessible objects are computed. Then the pair with the shortest distance is chosen. 

\subsubsection{Analysis of the algorithms}
\label{sec:analysis}

We provide proofs for time complexity and completeness of \textsc{Search} and \textsc{Greedy}. The optimality of the uniform-search is already proven~\cite{russell2002artificial}. Obviously, the time complexity of \textsc{Search} is exponential as it uses the uniform-cost search. %For those cases where an immediate relocation without planning is desirable, \textsc{Greedy} can replace. 
%However, \textsc{Search} has a reasonably short practical running time that will be shown empirically in Sec.~\ref{sec:exp}. 

\thm \textbf{4.1.} \textsc{Greedy} runs in polynomial time.

\pf. Given $\mathcal{O}_R$, checking if an edge exists between the robot node and the node of the first object in $\mathcal{O}_R$ is be done in $O(|E|) = O(N^2)$ without an adjacency matrix. If both robots can relocate the object, Euclidean distances between two pairs of locations are computed in constant time. If no robot can relocate the object, at most $2N$ computations are necessary to compute the distances between the robots and objects. The edge search and the distance computation repeat at most $|\mathcal{O}_R| \le N$ times. The overall time complexity is $O((N^2 + 2N)N) = O(N^3)$. \qed

\thm \textbf{4.2.} \textsc{Search} and \textsc{Greedy} are complete if the T-graph is connected.

\pf. In Sec.~\ref{sec:prob}, we assume that at least one object can be relocated (i.e., Assumption (iii)). The algorithms always return an object and the robot to relocate the object. If the robot fails to relocate the object, replanning returns another solution. In the worst case, all objects are relocated where the target object comes last.\qed
\smallskip

\subsection{Minimum makespan action sequencing}
\vspace{-2pt}

%Although the action sequencing problem for two manipulators is trivial, 
The most significant advantage of using multiple manipulators is certainly the capability to move simultaneously so multiple actions can be done in parallel. As illustrated in Fig.~\ref{fig:makespan}, we develop a simple method that parallelizes \texttt{pick} and \texttt{place} action primitives if the robots manipulate different objects. It inserts \texttt{standby} in between two consecutive \texttt{pick} and \texttt{place} (or vice versa) to let the robots start every action from a predefined pose. 

Once the action sequence is determined, the method performs two-arm motion planning for the parallelized actions. Even if the actions are not parallelized, two-arm motion planning is done to avoid conflicts between the robots. In order to take the most advantage out of two manipulators, we always assign a target state to the robot that is not being planned to perform \texttt{pick}. For \textsc{Search}, we precompute motions for all sequenced actions. With \textsc{Greedy}, motion planning is done after each robot-task pair is determined.

%Seems like we can't include the details due to the space limit...
%with successful motion plan and refer to the last successfully planned action of the arm to assign next non-picking target state. If one arm has a successful motion plan to a picking pose as the latest successful plan, the arm is assigned to place the picked object, as the other arm either moves to pick another object, or move to the standby state if it is not the next picking arm. Note that both standby state and all placing states are predefined prior to the task-planning. Robots may move directly from picking pose to a placing state, but we have them set to move to standby state in-between two consecutive actions for a practical purpose.

\section{Experiments}
\label{sec:exp}
\vspace{-2pt}

\begin{figure}
%\vspace{-5pt}
\captionsetup{skip=0pt}
    \captionsetup{skip=0pt}
    \centering
   \begin{subfigure}{0.235\textwidth}
   \captionsetup{skip=0pt}
	\includegraphics[width=\textwidth]{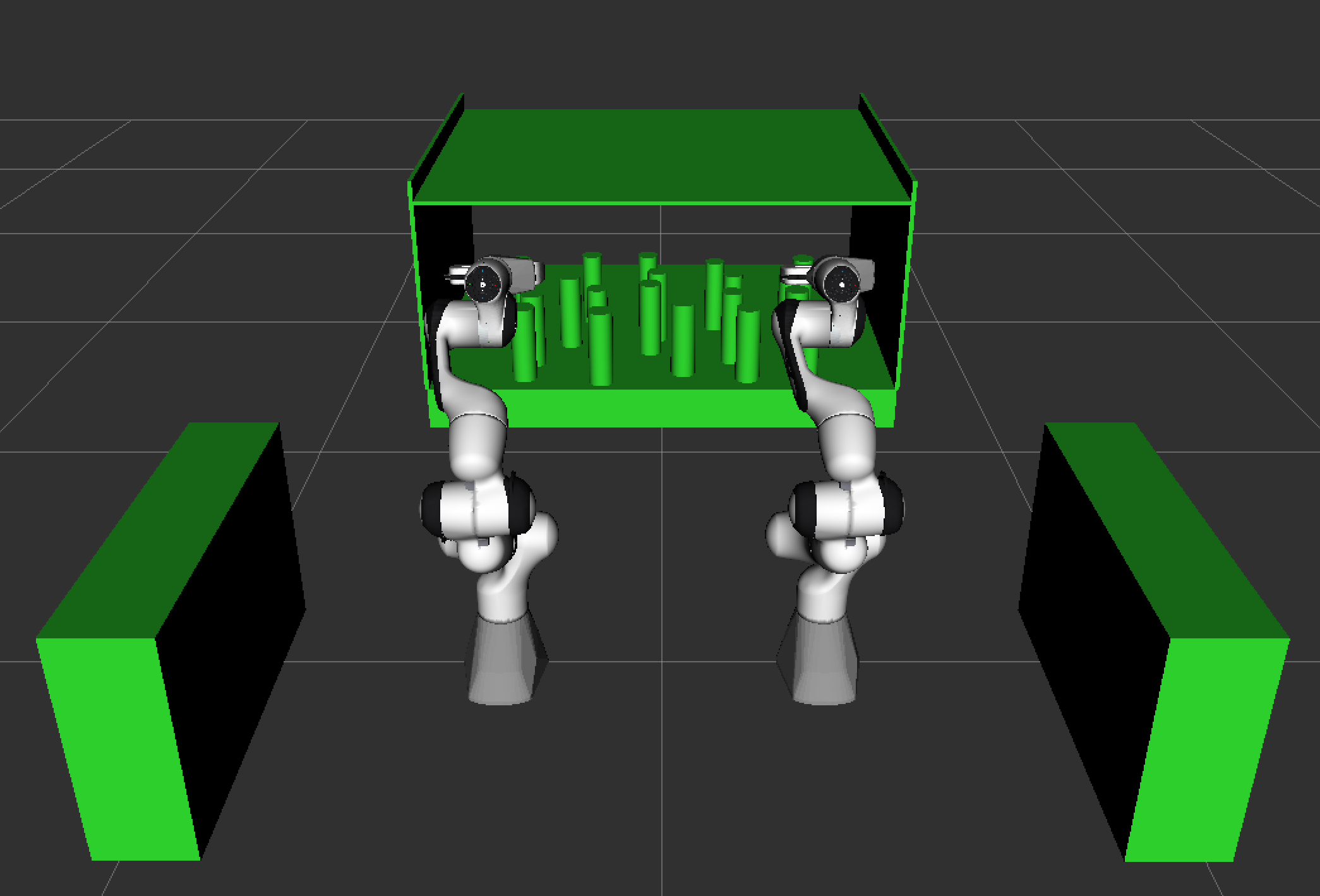}
	\caption{A MoveIt planning scene}
    \label{fig:moveit}
  \end{subfigure}
  \begin{subfigure}{0.235\textwidth}
  \captionsetup{skip=0pt}
	\includegraphics[width=\textwidth]{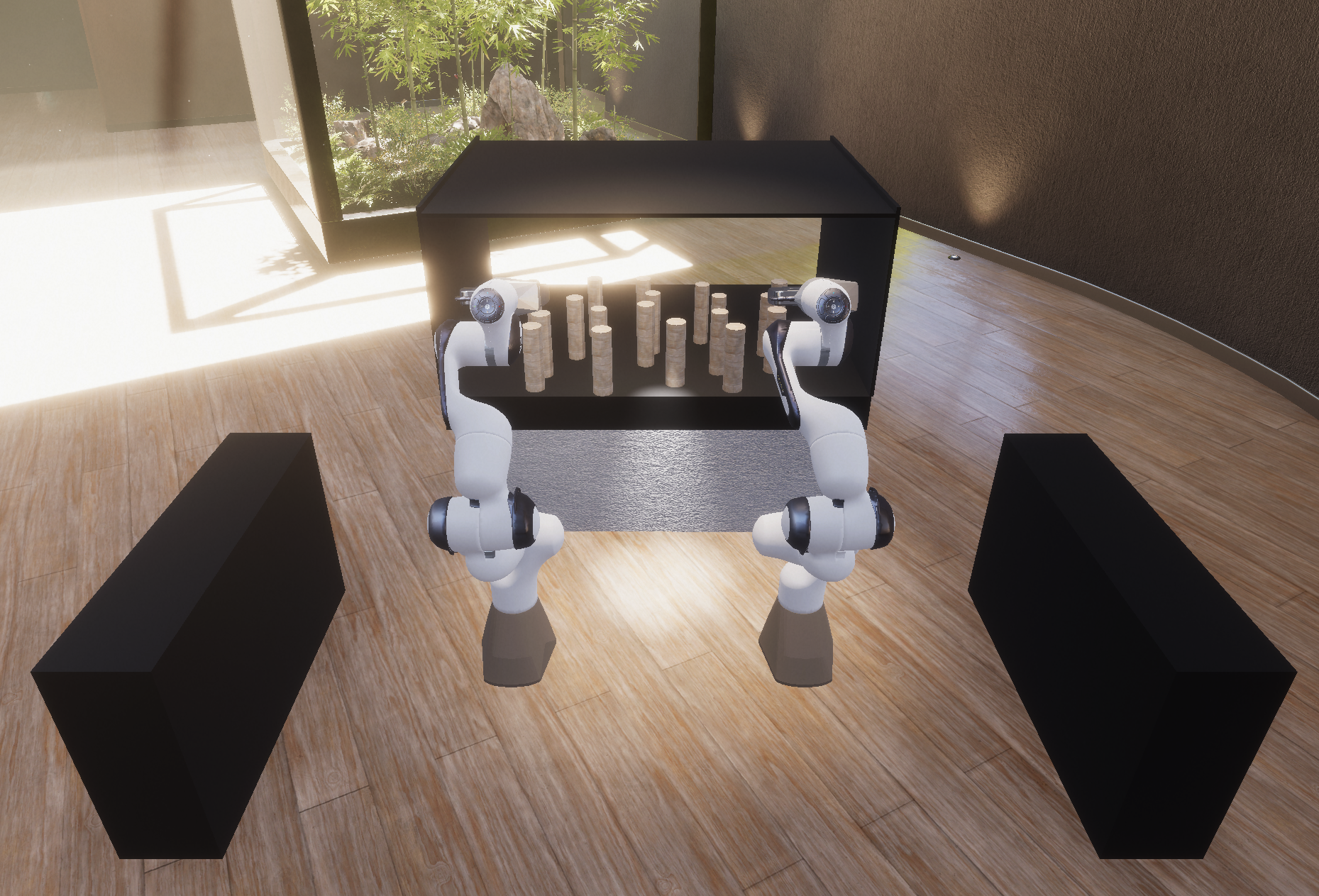}
	\caption{Unity simulation}
    \label{fig:unity}
  \end{subfigure}
  \caption{The environment for experiments. (a) MoveIt used for motion planning and simulations considering kinematics only, (b) Unity-ROS integration for dynamic simulations}
  \label{fig:env}\vspace{-10pt}
\end{figure}

We show the results from two sets of experiments. The first set evaluates \textsc{Search} and \textsc{Greedy} without dynamic simulations (Fig.~\ref{fig:moveit}). We measure the objective value, which is the number of turn-takings and the turn-taking rate. Since every solution could have a different number of tasks (so the number of chances for turn-taking varies), showing the rate enables comparisons. We also report other values related to performance like the numbers of relocated objects and replanning, the success rate within a time limit (200 seconds), and task and motion planning (TAMP) time. We have another set of experiments in a dynamic simulation environment (Fig.~\ref{fig:unity}) implemented using the Unity-ROS integration~\cite{unity}. In this set, we measure the execution time of the allocated tasks. We expect that the execution time decreases as the number of turn-taking increases. Since we are not aware of any comparable methods in the literature, we develop two simple but reasonable methods to compare:
\begin{itemize}
\item \textsc{Distance}: It allocates tasks according to the Euclidean distance between objects and robots if both robots can access the object to be relocated. Given $\mathcal{O}_R$, this method checks if the first object is accessible to both robots. If both robots can access the object, the robot whose end-effector is closer to the object is chosen. If only one robot can access the object, that robot is chosen. If no collision-free motion is found for both of the robots, $\mathcal{O}_R$ is updated for replanning. The choice repeats until the target is retrieved.
\item \textsc{Random}: It works similar to \textsc{Distance} but allocates tasks at random if both robots can access the object to be relocated.
\end{itemize}

%We expect that these compared two methods run fast but the quality of the solutions may be worse than our methods. 

We use RRTConnect~\cite{kuffner2000rrt} for motion planning implemented in the Open Motion Planning Library~\cite{sucan2012open} in MoveIt~\cite{moveit}. The system for the experiments is with AMD Ryzen 7 2700 3.2GHz with 32G RAM. The dimension of the confined space is 1100\,mm $\times$ 500\,mm $\times$ 850\,mm. The number of objects in the space is up to 20 so the space is cluttered with the objects and cramped for the bulky manipulators Franka Emika Panda with 7-DOF. 

\subsection{Task and motion planning}
\vspace{-2pt}

\begin{table*}
\captionsetup{skip=0pt}
\caption{Results of the proposed and compared methods (30 repetitions). Time is measured in seconds. The numbers in parentheses are standard deviation.}%
\label{tab:moveit_result}
\centering
\scalebox{0.76}{%
\begin{tabular}{|c||c|c|c|c|c|c|c|c|c|c|c|c|}
\hline
 \multirow{2}{*}{Measure} & \multicolumn{4}{c|}{$N = 12$} & \multicolumn{4}{c|}{$N = 16$}& \multicolumn{4}{c|}{$N = 20$}\\
\cline{2-13}
& \textsc{Search} & \textsc{Greedy} & \textsc{Distance} & \textsc{Random} &  \textsc{Search} & \textsc{Greedy} & \textsc{Distance} & \textsc{Random} &  \textsc{Search} & \textsc{Greedy} & \textsc{Distance} & \textsc{Random} \\
\hline
\#turn-taking & 1.13 (0.63) &  1.17 (1.02)  &  0.100 (0.31) &  0.633 (0.72)  & 1.2
 (1.03)  & 1.47 (1.50)  & 0.400 (0.86) & 1.17 (1.51)  & 1.63 (1.65) & 1.87 (2.03)  & 0.600(1.16) & 1.23 (1.33)   \\
\hline
Turn-taking rate (\%) & 82.2 (33.0) & 76.7 (43.0) & 6.67 (21.7) & 41.1 (45.9)  & 52.9 (37.7)  & 53.3 (44.1)  & 15.2 (31.4) & 39.5 (42.6) & 50.4 (37.4) & 42.8 (41.5) & 14.7 (30.3) & 33.1 (30.9)   \\
\hline
\#objects relocated & 2.40 (0.62) & 2.47 (0.82)  & 2.33 (0.61) & 2.73 (1.51)  & 3.30 (1.29)  & 3.87 (2.18)  & 3.63 (2.11) & 3.73 (2.21) & 4.00 (1.17) & 4.63 (2.09) & 4.40 (1.69) & 4.47 (1.83)   \\
\hline
Success rate (\%) & 100 & 100 & 100 & 100 & 100 & 93.3 & 96.7 & 93.3 & 93.3 & 93.3 & 96.7 & 96.7 \\
\hline
\#replanning & 0.333 & 0.267 & 0.100 & 0.733 & 0.300 & 0.933 & 0.567 & 0.733 & 0.0667 & 0.933 & 0.633 & 0.800\\
%Failing instances include those that exceed the time limit (3 minutes for task and motion planning (not including execution)
\hline
TAMP time  & 33.1 (16.3) & 24.1 (15.9) & 20.0 (9.18) & 30.7 (34.7) & 57.0 (29.3) & 48.0 (32.4) & 39.8 (23.6) & 42.3 (24.7) & 84.0 (26.4) & 76.2 (42.3) & 66.2 (33.2) & 69.9 (37.9)    \\
\hline
TAMP time per task  & 13.4 (4.64) & 9.26 (2.95) & 8.39 (1.93) & 9.59 (4.04) & 16.9 (2.81) & 12.2 (3.50) & 10.8 (2.25) & 11.3 (2.65) & 21.0 (2.40) & 15.9 (2.59) & 14.7 (2.13) & 15.1 (2.45)   \\
\hline
\end{tabular}}
\vspace{-10pt}
\end{table*}

We test the four methods with 30 instances for each $N = 12, 16, 20$ where the object locations are sampled uniformly at random. For fair comparisons, the set of test instances is identical across all methods. We also report the task and motion planning time for each task in order to assess the average time necessary to generate a plan for each task. 

\begin{figure}
%\vspace{-5pt}
    \captionsetup{skip=0pt}
    \centering
   \begin{subfigure}{0.222\textwidth}
   \captionsetup{skip=0pt}
	\includegraphics[width=\textwidth]{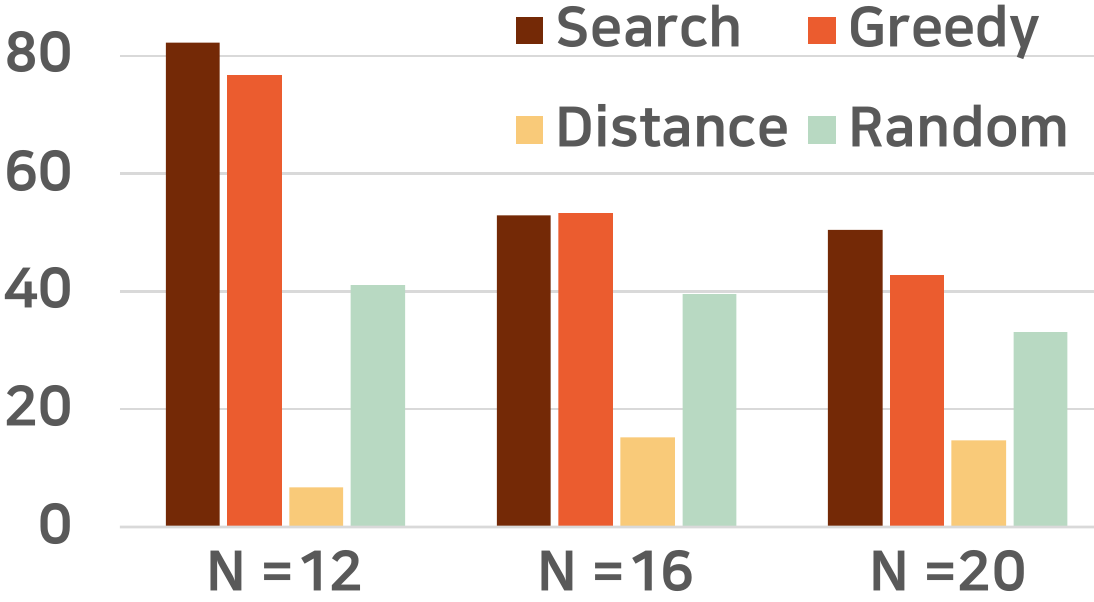}
	\caption{Turn-taking rate (\%)}
    %\caption{No collision while bringing $o_t$ to $(x_2, y_2)$}
    \label{fig:moveit_result_rate}
  \end{subfigure}%
  \begin{subfigure}{0.263\textwidth}
  \captionsetup{skip=0pt}
	\includegraphics[width=\textwidth]{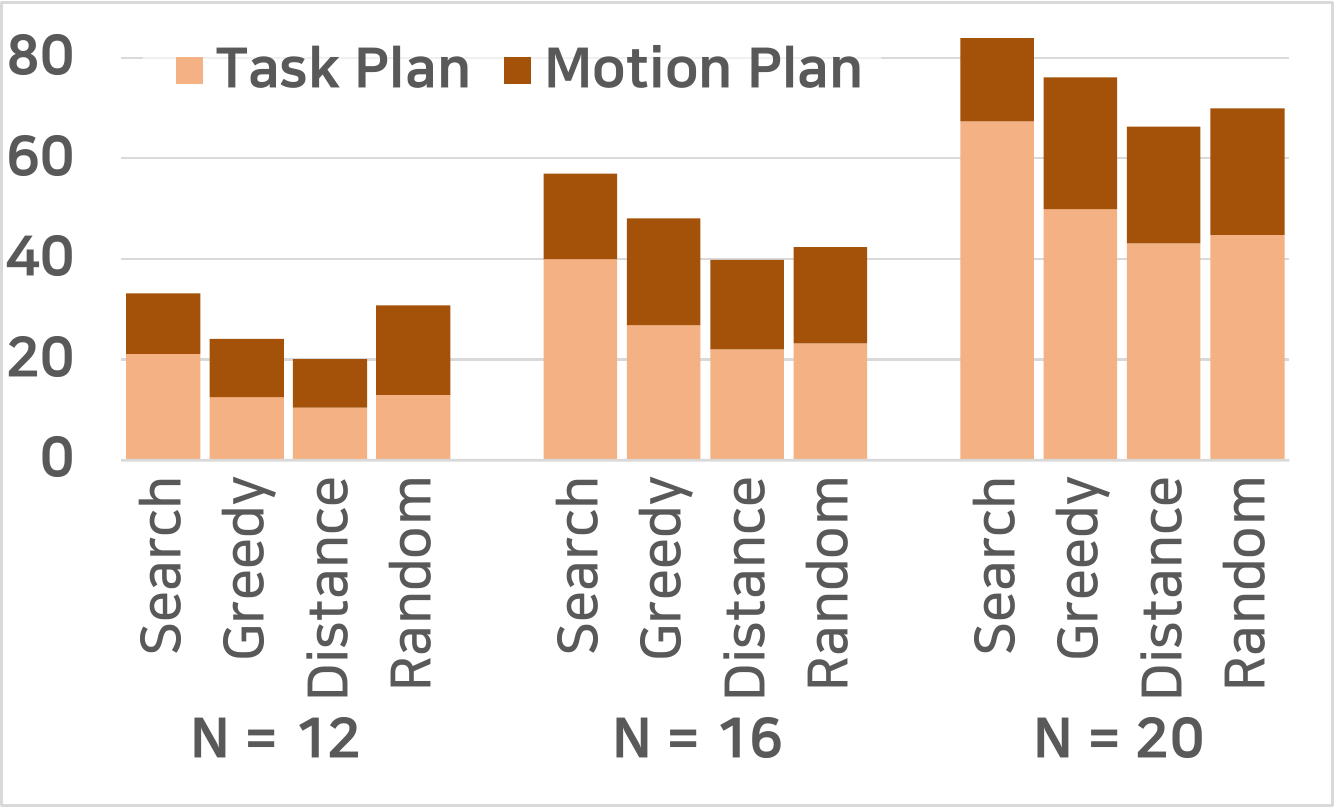}
	\caption{Task and motion planning time (sec)}
    %\caption{No collision while bringing $o_t$ and $o_2$ to $(x_5, y_5)$}\vspace{8pt}
    \label{fig:moveit_result_time}
  \end{subfigure}
  \caption{Important results from the experiment of task and motion planning}
  \label{fig:moveit_result}\vspace{-5pt}
\end{figure}

As shown in Table~\ref{tab:moveit_result} and Fig.~\ref{fig:moveit_result}, our proposed methods notably outperform in finding a solution with more turn-takings. With $N=12$, the turn-taking rate of \textsc{Search} is 82.2\%, which means that the robots can take turns more than four times out of five chances. An example allocation with this rate could be $\mathcal{X}_R = (r_1, r_2, r_1, r_2, r_1, r_1)$. For the most cluttered instance set ($N=20$), \textsc{Search} reduces the turn-taking rate at most 70.8\% and at least 34.3\% compared to \textsc{Distance} and \textsc{Random}, respectively. As a consequence, more turn-takings will lead to a more time-efficient task plan for the robots to execute because more actions can be parallelized, resulting in a shorter makespan. \textsc{Greedy} also shows high turn-taking rates than the compared results, although it does not outperform \textsc{Search}. Notice that the number of turn-takings (not the rate) of \textsc{Greedy} is larger than \textsc{Search}. This result does not contradict the fact that \textsc{Search} is optimal because the chances of turn-taking are greater with \textsc{Greedy} as it relocates more objects on average. This explains why we should focus on the turn-taking rate.

Despite the exponential time complexity of \textsc{Search}, its practical planning time (TAMP time in the table) is not prohibitively long. In comparison with others (shown in Fig.~\ref{fig:moveit_result_time}), its planning time is at most 21.2\% longer (when $N=20$), which is about 18 seconds longer. One reason for the bounded planning time is that each node in the search tree has a branching factor of two (i.e., has at most two children) since each branching is for each robot performing a task. Also, the number of objects to be relocated does not exceed four in average with \textsc{Search} so the search depth is also bounded. Therefore, \textsc{Search} can run efficiently even with its poor theoretical time complexity. The breakdowns in Fig.~\ref{fig:moveit_result_time} indicate that motion planning time is the shortest with \textsc{Search}. The main reason is that \textsc{Search} does not request frequent replannings leading not to generate many motion planning queries, which are computationally expensive. The success rate measures the chance of returning a solution within the time limit (200 seconds). All methods show high success rates. The last measure to note is the number of relocated objects where \textsc{Search} outperforms others in general. \textsc{Greedy} runs faster than \textsc{Search}, but the difference in time is not impressive since \textsc{Search} does not replan much in our experiments. If the environment is more dynamic so replanning occurs frequently, the difference will become significant.

\subsection{Dynamic simulations}
\vspace{-2pt}

\begin{figure}
% \vspace{-5pt}
    \captionsetup{skip=0pt}
    \centering
   \begin{subfigure}{0.239\textwidth}
   \captionsetup{skip=0pt}
	\includegraphics[width=\textwidth]{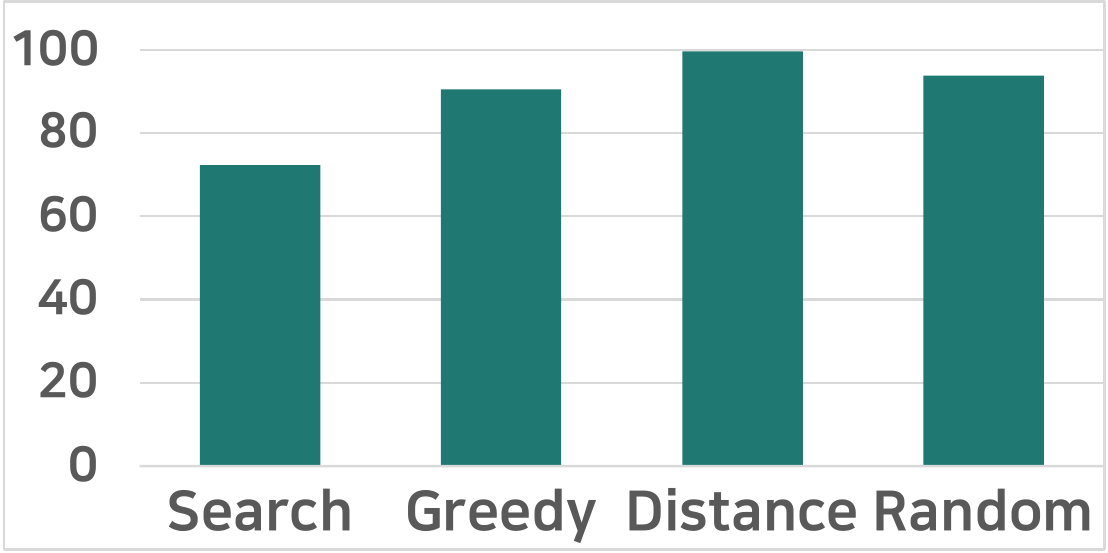}
	\caption{Execution time (sec)}
    \label{fig:time_unity_result}
  \end{subfigure}
  \begin{subfigure}{0.235\textwidth}
  \captionsetup{skip=0pt}
	\includegraphics[width=\textwidth]{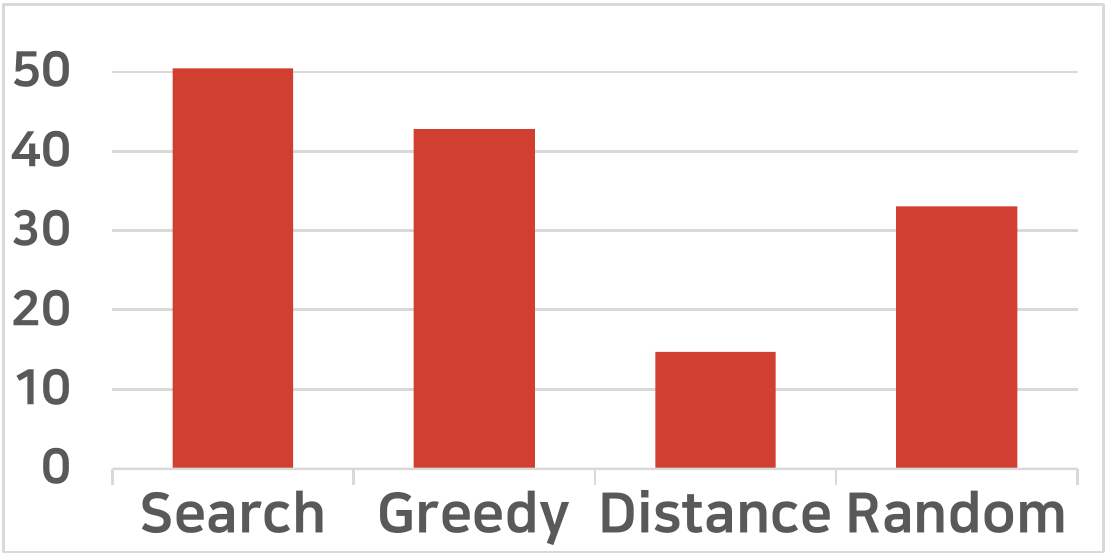}
	\caption{Turn-taking rate (\%)}
    \label{fig:object_unity_result}
  \end{subfigure}
  \caption{Important results from dynamic simulations}
  \label{fig:unity_result} \vspace{-10pt}
\end{figure}

In this set of experiments, we run the relocation plan in the dynamic simulation environment to show the feasibility of running our methods with real robots. The result from 30 instances where $N=20$ is shown in Table~\ref{tab:unity_result} and Fig.~\ref{fig:unity_result}. As expected, \textsc{Search} shows the shortest execution time with the highest number of turn-takings. It indicates that \textsc{Search} is capable of generating efficient plans for execution at runtime. The reduction in execution time is at most 27.3\% and at least 22.9\% compared to \textsc{Distance} and \textsc{Random}, respectively. Although it is not satisfactory, \textsc{Greedy} shows a slightly better performance. The low reduction in execution time (up to 9\%) comes from the myopic decision-making in contrast to the far-sighted \textsc{Search}.

%of manipulators. Though it is more time-expensive at the level of task planning, it remains the best as it performs significantly better in optimizing motion planner usage as well as planning a time-efficient task plans with notably higher rate of turn-takings of manipulators; it shows over 22\% more efficient in time compared to \textsc{Random} method and over 27\% more efficient in time compared to \textsc{Distance} method in average. Better performance in execute-time is specifically significant since it may scale up as the target velocity of motions is adjusted to be lowered for stabilization purposes in real world while planning times are not dependent on the velocity.

% Do we really do this??? Nah
%We also show the result of performing the relocation plan by one robot (\textsc{Single}) to see (i) how much planning overhead is caused by coordinating two robots and (ii) how much completion time can be saved with the simultaneous execution of tasks. The result is shown in Table~\ref{tab:unity_result}. (ADD MORE)

\begin{table}
\captionsetup{skip=0pt}
\caption{Results of dynamic simulation in the Unity-ROS environment (30 repetitions and $N=20$)}
\label{tab:unity_result}
\centering
\scalebox{0.88}{%
\begin{tabular}{|c||c|c|c|c|}
\hline
Measure & \textsc{Search} & \textsc{Greedy} & \textsc{Distance} & \textsc{Random} \\
\hline
\#turn-taking & 1.63 (1.65) & 1.87 (2.03) & 0.600 (1.163)  & 1.23 (1.33)\\
\hline
Turn-taking rate (\%) & 50.4 (37.4) & 42.8 (41.5) & 14.7 (30.3)  & 33.1 (30.9)\\
\hline
\#objects relocated & 4.00 (1.17) & 4.63 (2.09) & 4.40 (1.69)  & 4.47 (1.83)\\
\hline
Success rate (\%) & 93.3 & 93.3 & 96.7 & 96.7  \\
\hline
\#replanning & 0.0667 & 0.933 & 0.633 & 0.800 \\
%Maybe include the failing instances that exceed 3 mins of the time limit for task and motion planning (not execution)
\hline
TAMP time & 84.0 (26.4) & 76.2 (42.3) & 66.2 (33.2)  & 69.9 (37.9)\\
\hline
Total execution time & 72.4 (24.3) & 90.6 (46.4) & 99.6 (47.3)  & 93.9 (45.7) \\
\hline
\end{tabular}}
%\vspace{-15pt}
\end{table}

\section{Conclusion}
\vspace{-2pt}

We consider the problem of coordinating two manipulators to retrieve a target object from a cluttered and confined space where overhand grasps are not allowed. The objective is to minimize the execution time of the relocation plan through simultaneous executions of tasks by the two robots. We formulate the problem as three subproblems: (A) computing relocation tasks, (B) allocating the tasks to the robots, and (C) sequencing primitive actions and generating collision-free motions for the actions. We present an approach that solves the three subproblems sequentially. We propose two methods where one computes the entire task plan and allocation before execution, and the other finds a relocation task and the allocated robot for the task instantaneously after executing every task. We provide proofs for the time complexities and completeness of the methods. Our extensive experiments show that our methods significantly reduce the execution time up to 27.3\% compared to other methods that do not consider coordination of the robots. In the future, we will work on developing a decentralized method for coordination and motion planning. Also, we will run the proposed methods on more than two robots to show their scalability.

%asynchronous motion scheduling, velocity tuning, 
%Fully distribute: task allocation, motion scheduling
%N arms
%different shapes (t-ro paper)
%occlusion (partially handled in this paper but refer to t-ro)
%Prediction of object states but not just instant replanning with updated object states

\bibliographystyle{IEEEtran}
\bibliography{references}

\end{document}